\documentclass{article}
\usepackage{ijcai24}
\usepackage{times}
\usepackage{soul}
\usepackage{url}
\usepackage[hidelinks]{hyperref}
\usepackage[utf8]{inputenc}
\usepackage[small]{caption}
\usepackage{graphicx}
\usepackage{amsmath}
\usepackage{amsthm}
\usepackage{booktabs}
\usepackage{algorithm}
\usepackage{algorithmic}
\usepackage[switch]{lineno}
\usepackage{color}
\usepackage{multirow}
\usepackage{bbding}
\usepackage{pifont}
\usepackage{wasysym}
\usepackage{amssymb}
\usepackage{amsfonts,amssymb}
\usepackage{colortbl}
\definecolor{mygray}{gray}{.9}

\usepackage{titlesec}

\title{Unified Physical-Digital Face Attack Detection}

\author{
	Hao Fang$^{\rm 1}$\thanks{These authors contributed equally to this work},
	Ajian Liu$^{\rm 1*}$,
	Haocheng Yuan$^{\rm 2}$,
	Junze Zheng$^{\rm 2}$,
	Dingheng Zeng$^{\rm 3}$,
	Yanhong Liu$^{\rm 3}$
	Jiankang Deng$^{\rm 4}$,
	Sergio Escalera$^{\rm 5}$,
	Xiaoming Liu$^{\rm 6}$,
	Jun Wan$^{\rm 1,2}$\thanks{Corresponding author},
	Zhen Lei$^{\rm 1}$ \\
	$^{\rm 1}$MAIS, CASIA, China; 
	$^{\rm 2}$MUST, China; 
	$^{\rm 3}$Mashang Consumer Finance Co., Ltd., China \\
	$^{\rm 4}$Imperial College London, UK; 
	$^{\rm 5}$Computer Vision Center, Spain; 
	$^{\rm 6}$MSU, USA
	\tt\footnotesize
	\{fanghao2021,ajian.liu,jun.wan\}@ia.ac.cn, 
	\tt\footnotesize
}

\begin{document}

\maketitle
\begin{abstract}
      Face Recognition (FR) systems can suffer from physical (i.e., print photo) and digital (i.e., DeepFake) attacks. However, previous related work rarely considers both situations at the same time. This implies the deployment of multiple models and thus more computational burden. The main reasons for this lack of an integrated model are caused by two factors: (1) The lack of a dataset including both physical and digital attacks with ID consistency which means the same ID covers the real face and all attack types; (2) Given the large intra-class variance between these two attacks, it is difficult to learn a compact feature space to detect both attacks simultaneously. To address these issues, we collect a Unified physical-digital Attack dataset, called \textbf{UniAttackData}. 
      The dataset consists of $1,800$ participations of $2$ and $12$ physical and digital attacks, respectively, resulting in a total of $29,706$ videos. 
      Then, we propose a Unified Attack Detection framework based on Vision-Language Models (VLMs), namely \textbf{UniAttackDetection}, which includes three main modules: the Teacher-Student Prompts (TSP) module, focused on acquiring unified and specific knowledge respectively; the Unified Knowledge Mining (UKM) module, designed to capture a comprehensive feature space; and the Sample-Level Prompt Interaction (SLPI) module, aimed at grasping sample-level semantics. These three modules seamlessly form a robust unified attack detection framework. Extensive experiments on UniAttackData and three other datasets demonstrate the superiority of our approach for unified face attack detection. 
\end{abstract}

\section{Introduction}
\begin{figure}[t]
\centering
\includegraphics[width=1.0\linewidth]{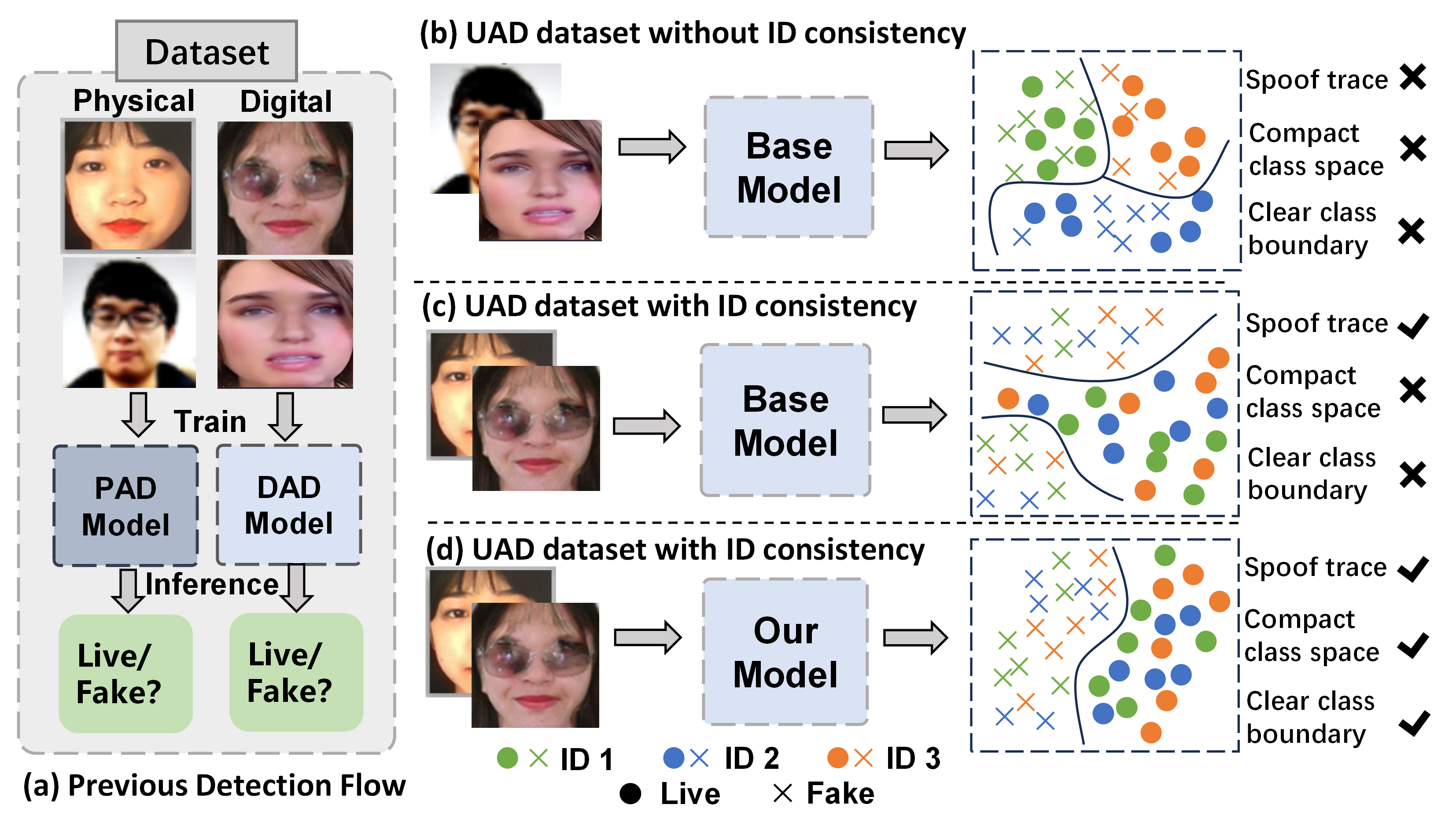}
\caption{\textbf{Paradigm Comparison.} (a) Prior approaches necessitate the separate training and deployment of PAD and DAD models, demanding significant computational resources and inference time. (b) UAD dataset without ID consistency introduces the risk of the algorithm learning noise related to ID. (c) The base model encounters challenges in acquiring a compact feature space when confronted with the UAD dataset. (d) Our algorithm learns compact feature space and clear class boundaries.}
\label{fig:spoofing-fake2.0}
\end{figure}

Face recognition (FR) system has been widely used in face unlocking, face payment, and video surveillance. 
It can face a diverse set of attacks: (1) Physical attacks (PAs), 
i.e. print-attack, replay-attack and mask-attack~\cite{liu2019multi,liu2022contrastive,fang2023surveillance}; and (2) Digital attacks (DAs). 
Both Physical Attack Detection (PAD)~\cite{Liu2018Learning,liu2021face,Liu2023FMViTFM,ijcai2022p165} and Digital Attack Detection (DAD)~\cite{dang2020detection,zhao2021multi} are still being studied by related works as two independent tasks. 
As shown in Fig.~\ref{fig:spoofing-fake2.0}(a), this will lead to the training of both physical and digital attack detection models and their deployment, requiring large computing resources which increases the inference time.

Two main reasons hinder the unification of detecting physical-digital attacks: (1) \textbf{Lack of the physical-digital dataset.} Although two datasets (namely GrandFake~\cite{deb2023unified} and JFSFDB~\cite{yu2022benchmarking}) are proposed to tackle this problem, they only simply merge PAs and DAs datasets. They cannot guarantee each ID covers the real face and all attack types, which would lead the model to learn the liveness-irrelevant signals such as face ID, domain-specific information, and background. Therefore, the lack of a physical-digital dataset with ID consistency limits the research on unified detection algorithms. 
\textbf{(2) Distinct intra-class variances.} Although both physical and digital attacks are classified as fake in the solution space of the face anti-spoofing problem, the vast differences between these two types of attacks increase the intra-class distances. Most of the existing attack detection methods~\cite{Liu2018Learning,yu2020searching,liu2022disentangling} are proposed for a specific attack, thus overfitting to a certain attack and failing to learn a compact feature space to detect both attacks simultaneously.
To solve the first issue, we collect and release a dataset combining physical and digital attacks, named UniAttackData, which contains $2$ physical and $12$ digital attacks for each of $1,800$ subjects, with a total of $29,706$ videos. Compared to GrandFake and JFSFDB datasets, the UniAttackData provides four advantages: (1) \textbf{The most complete attack types for each face ID.} Our dataset constructs physical-digital attacks for each face ID, rather than simply merging existing PAs and DAs datasets. 
(2) \textbf{The most advanced forgery method.} For digital attacks, we select the most advanced digital forgery methods in the past three years. For physical attacks, we consider printing attacks and video replay attacks in various environments and physical media. 
(3) \textbf{The most amount of images.} The proposed dataset is the largest one in terms of the number of images, which is more than $3.2 \times$ boosted compared to the previous face anti-spoofing dataset like GrandFake. 
(4) \textbf{The most diverse evaluation protocols.} Beyond the within-modal evaluation protocols, we also provide the cross-attack evaluation protocols in our dataset, in which algorithms trained in one attack category are evaluated in other attack categories. 
For the second issue, we propose a unified detection framework, named UniAttackDetection, which has three main modules: \textbf{(1) Introducing textual information to depict visual concepts.} Learning joint and category-specific knowledge features by constructing teacher prompts and student prompts, respectively. \textbf{(2) Enhanced learning of complete feature space.} We propose joint knowledge mining loss to learn complete and tight feature spaces using text-to-text optimization. \textbf{(3) Sample-level visual text interaction.} We map learnable text tokens to the visual embedding space and understand sample-level semantics through multi-modal prompt learning. To sum up, the main contributions of this paper are summarized as follows:
\begin{itemize}
	\setlength{\itemsep}{1.0pt}
	\item
	We propose a dataset that combines physical and digital attacks called UniAttackData. To the best of our knowledge, this is the first unified attack dataset with guaranteed ID consistency.
	\item
	We propose a unified attack detection framework based on the vision language model, named UniAttackDetection. It adaptively learns a tight and complete feature space with the help of extensive visual concepts and rich semantic information in text prompts.
	\item
	We conducted extensive experiments on UniAttackData and three existing attack datasets. The results show the superiority of our approach in the task of unified face attack detection.
\end{itemize}

\section{Related Work}
\subsection{Physical and Digital Attack Detection}
PAD technology aims to identify whether the face collected by sensors comes from a live face or is a presentation attack. 
The most advanced algorithms~\cite{Liu2018Learning,yu2020nasfas,Liu_2023_CVPR} are based on facial depth to determine authenticity. However, their performance severely deteriorates in unknown domains. At present, the generalization of FAS algorithm~\cite{Rui2019Multi} is increasingly becoming an important evaluation indicator. Digital attack detection aims to distinguish authentic facial images from digitally manipulated facial artifacts. In initial studies~\cite{rossler2019faceforensics++}, image classification backbones (such as Xception~\cite{chollet2017xception}) were employed to extract features from isolated facial images, facilitating binary classification. With the increasing visual realism of forged faces, recent efforts~\cite{zhao2021multi,frank2020leveraging} focus on identifying more reliable forgery patterns, including noise statistics, local textures, and frequency information. 

\begin{table*}[ht]
  \scalebox{0.57}{
    \begin{tabular}{|c|c|cl|c|cc|ccc|}
    \hline
    \multirow{3}{*}{\textit{Dataset}} &
    \multirow{3}{*}{Attack Type (each ID)} &
    \multicolumn{2}{c|}{\multirow{3}{*}{\# Datasets / Data}} &
    \multirow{3}{*}{\# ID} &
    \multicolumn{2}{c|}{\multirow{2}{*}{Physical Attacks}} &
    \multicolumn{3}{c|}{\multirow{2}{*}{Digital Attacks}} \\
    &
    &
    \multicolumn{2}{c|}{} &
    &
    \multicolumn{2}{c|}{} &
    \multicolumn{3}{c|}{} \\ \cline{6-10}
      &
      &
      \multicolumn{2}{c|}{} &
      &
      \multicolumn{1}{c|}{Dataset Name} &
      No. &
      \multicolumn{1}{c|}{\# Categories} &
      \multicolumn{1}{c|}{Methods} &
      No. \\ \hline
      \multirow{12}{*}{\textit{GrandFake}} &
      \multirow{12}{*}{Incomplete} &
      \multicolumn{2}{c|}{\multirow{12}{*}{\begin{tabular}[c]{@{}c@{}}6 sets:\\      789412 (I)\\    (Live: 341738, Fake: 447674)\end{tabular}}} &
      \multirow{12}{*}{96817} &
      \multicolumn{1}{c|}{\multirow{12}{*}{SiW-M ~\cite{liu2019deep}}} &
      \multirow{12}{*}{128112 (I)} &
      \multicolumn{1}{c|}{\multirow{6}{*}{\begin{tabular}[c]{@{}c@{}}Adv\\      (6)\end{tabular}}} &
      \multicolumn{1}{c|}{FGSM ~\cite{goodfellow2015explaining}} &
      19739   (I) \\ \cline{9-10} 
      &
      &
      \multicolumn{2}{c|}{} &
      &
      \multicolumn{1}{c|}{} &
      &
      \multicolumn{1}{c|}{} &
      \multicolumn{1}{c|}{PGD ~\cite{madry2019deep}} &
      19739   (I) \\ \cline{9-10} 
      &
      &
      \multicolumn{2}{c|}{} &
      &
      \multicolumn{1}{c|}{} &
      &
      \multicolumn{1}{c|}{} &
      \multicolumn{1}{c|}{DeepFool ~\cite{moosavidezfooli2016deepfool}} &
      19739   (I) \\ \cline{9-10} 
      &
      &
      \multicolumn{2}{c|}{} &
      &
      \multicolumn{1}{c|}{} &
      &
      \multicolumn{1}{c|}{} &
      \multicolumn{1}{c|}{AdvFaces ~\cite{deb2019advfaces}} &
      19739   (I) \\ \cline{9-10} 
      &
      &
      \multicolumn{2}{c|}{} &
      &
      \multicolumn{1}{c|}{} &
      &
      \multicolumn{1}{c|}{} &
      \multicolumn{1}{c|}{GFLM ~\cite{dabouei2018fast}} &
      17946   (I) \\ \cline{9-10} 
      &
      &
      \multicolumn{2}{c|}{} &
      &
      \multicolumn{1}{c|}{} &
      &
      \multicolumn{1}{c|}{} &
      \multicolumn{1}{c|}{SemanticAdv ~\cite{qiu2020semanticadv}} &
      19739   (I) \\ \cline{8-10} 
      &
      &
      \multicolumn{2}{c|}{} &
      &
      \multicolumn{1}{c|}{} &
      &
      \multicolumn{1}{c|}{\multirow{6}{*}{\begin{tabular}[c]{@{}c@{}}DeepFake\\      (6)\end{tabular}}} &
      \multicolumn{1}{c|}{FaceSwap ~\cite{faceswap}} &
      14492   (I) \\ \cline{9-10} 
      &
      &
      \multicolumn{2}{c|}{} &
      &
      \multicolumn{1}{c|}{} &
      &
      \multicolumn{1}{c|}{} &
      \multicolumn{1}{c|}{Deepfake ~\cite{korshunov2018deepfakes}} &
      18165   (I) \\ \cline{9-10} 
      &
      &
      \multicolumn{2}{c|}{} &
      &
      \multicolumn{1}{c|}{} &
      &
      \multicolumn{1}{c|}{} &
      \multicolumn{1}{c|}{Face2Face ~\cite{thies2016face2face}} &
      18204   (I) \\ \cline{9-10} 
      &
      &
      \multicolumn{2}{c|}{} &
      &
      \multicolumn{1}{c|}{} &
      &
      \multicolumn{1}{c|}{} &
      \multicolumn{1}{c|}{StarGAN ~\cite{choi2018stargan}} &
      45473   (I) \\ \cline{9-10} 
      &
      &
      \multicolumn{2}{c|}{} &
      &
      \multicolumn{1}{c|}{} &
      &
      \multicolumn{1}{c|}{} &
      \multicolumn{1}{c|}{STGAN ~\cite{liu2019stgan}} &
      29983   (I) \\ \cline{9-10} 
      &
      &
      \multicolumn{2}{c|}{} &
      &
      \multicolumn{1}{c|}{} &
      &
      \multicolumn{1}{c|}{} &
      \multicolumn{1}{c|}{StyleGAN2 ~\cite{Karras2019stylegan2}} &
      76604   (I) \\ \hline
      \multirow{6}{*}{\textit{JFSFDB}} &
      \multirow{6}{*}{Incomplete} &
      \multicolumn{2}{c|}{\multirow{6}{*}{\begin{tabular}[c]{@{}c@{}}9 sets:\\      27172 (V)\\(Live: 5650, Fake: 21522)\end{tabular}}} &
      \multirow{6}{*}{356} &
      \multicolumn{1}{c|}{SiW ~\cite{Liu2018Learning}} &
      3173 (V) &
      \multicolumn{1}{c|}{\multirow{6}{*}{\begin{tabular}[c]{@{}c@{}}DeepFake\\      (4)\end{tabular}}} &
      \multicolumn{1}{c|}{Face2Face ~\cite{thies2016face2face}} &
      1000  (V) \\ \cline{6-7} \cline{9-10} 
      &
      &
      \multicolumn{2}{c|}{} &
      &
      \multicolumn{1}{c|}{3DMAD ~\cite{erdogmus2014spoofing}} &
      85 (V) &
      \multicolumn{1}{c|}{} &
      \multicolumn{1}{c|}{FaceSwap ~\cite{faceswap}} &
      1000  (V) \\ \cline{6-7} \cline{9-10} 
      &
      &
      \multicolumn{2}{c|}{} &
      &
      \multicolumn{1}{c|}{HKBU ~\cite{liu20163d}} &
      588 (V) &
      \multicolumn{1}{c|}{} &
      \multicolumn{1}{c|}{\multirow{2}{*}{NeuralTextures ~\cite{thies2019deferred}}} &
      \multirow{2}{*}{1000  (V)} \\ \cline{6-7}
      &
      &
      \multicolumn{2}{c|}{} &
      &
      \multicolumn{1}{c|}{MSU ~\cite{wen2015face}} &
      210 (V) &
      \multicolumn{1}{c|}{} &
      \multicolumn{1}{c|}{} &
      \\ \cline{6-7} \cline{9-10} 
      &
      &
      \multicolumn{2}{c|}{} &
      &
      \multicolumn{1}{c|}{3DMask ~\cite{yu2020nasfas}} &
      864 (V) &
      \multicolumn{1}{c|}{} &
      \multicolumn{1}{c|}{\multirow{2}{*}{Deepfake ~\cite{korshunov2018deepfakes}}} &
      \multirow{2}{*}{10752  (V)} \\ \cline{6-7}
      &
      &
      \multicolumn{2}{c|}{} &
      &
      \multicolumn{1}{c|}{ROSE ~\cite{li2018unsupervised}} &
      2850 (V) &
      \multicolumn{1}{c|}{} &
      \multicolumn{1}{c|}{} &
      \\ \hline
      \multirow{14}{*}{\textit{\textbf{\begin{tabular}[c]{@{}c@{}}UniAttackData\\      (Ours)\end{tabular}}}} &
      \multirow{14}{*}{\textbf{Complete}} &
      \multicolumn{2}{c|}{\multirow{14}{*}{\textbf{\begin{tabular}[c]{@{}c@{}}1 set:\\      29706 (V)\\(Live: 1800, Fake: 27906)\end{tabular}}}} &
      \multirow{14}{*}{\textbf{1800}} &
      \multicolumn{1}{c|}{\multirow{14}{*}{\begin{tabular}[c]{@{}c@{}}CASIA-SURF\\      CeFA ~\cite{liu2021casia}\end{tabular}}} &
      \multirow{14}{*}{6400 (V)} &
      \multicolumn{1}{c|}{\multirow{6}{*}{\begin{tabular}[c]{@{}c@{}}Adv\\      (6)\end{tabular}}} &
      \multicolumn{1}{c|}{advdrop ~\cite{duan2021advdrop}} &
      1706 (V) \\ \cline{9-10} 
      &
      &
      \multicolumn{2}{c|}{} &
      &
      \multicolumn{1}{c|}{} &
      &
      \multicolumn{1}{c|}{} &
      \multicolumn{1}{c|}{alma ~\cite{rony2021augmented}} &
      1800 (V) \\ \cline{9-10} 
      &
      &
      \multicolumn{2}{c|}{} &
      &
      \multicolumn{1}{c|}{} &
      &
      \multicolumn{1}{c|}{} &
      \multicolumn{1}{c|}{demiguise ~\cite{wang2021demiguise}} &
      1800 (V) \\ \cline{9-10} 
      &
      &
      \multicolumn{2}{c|}{} &
      &
      \multicolumn{1}{c|}{} &
      &
      \multicolumn{1}{c|}{} &
      \multicolumn{1}{c|}{fgtm ~\cite{zou2022making}} &
      1800 (V) \\ \cline{9-10} 
      &
      &
      \multicolumn{2}{c|}{} &
      &
      \multicolumn{1}{c|}{} &
      &
      \multicolumn{1}{c|}{} &
      \multicolumn{1}{c|}{ila\_da ~\cite{yan2022ila}} &
      1800 (V) \\ \cline{9-10} 
      &
      &
      \multicolumn{2}{c|}{} &
      &
      \multicolumn{1}{c|}{} &
      &
      \multicolumn{1}{c|}{} &
      \multicolumn{1}{c|}{ssah ~\cite{luo2022frequency}} &
      1800 (V) \\ \cline{8-10} 
      &
      &
      \multicolumn{2}{c|}{} &
      &
      \multicolumn{1}{c|}{} &
      &
      \multicolumn{1}{c|}{\multirow{6}{*}{\begin{tabular}[c]{@{}c@{}}DeepFake\\      (6)\end{tabular}}} &
      \multicolumn{1}{c|}{FaceDancer ~\cite{rosberg2023facedancer}} &
      1800 (V) \\ \cline{9-10} 
      &
      &
      \multicolumn{2}{c|}{} &
      &
      \multicolumn{1}{c|}{} &
      &
      \multicolumn{1}{c|}{} &
      \multicolumn{1}{c|}{InsightFace ~\cite{2020Deep}} &
      1800 (V) \\ \cline{9-10} 
      &
      &
      \multicolumn{2}{c|}{} &
      &
      \multicolumn{1}{c|}{} &
      &
      \multicolumn{1}{c|}{} &
      \multicolumn{1}{c|}{SimSwap ~\cite{chen2020simswap}} &
      1800 (V) \\ \cline{9-10} 
      &
      &
      \multicolumn{2}{c|}{} &
      &
      \multicolumn{1}{c|}{} &
      &
      \multicolumn{1}{c|}{} &
      \multicolumn{1}{c|}{SAFA ~\cite{wang2021safa}} &
      1800 (V) \\ \cline{9-10} 
      &
      &
      \multicolumn{2}{c|}{} &
      &
      \multicolumn{1}{c|}{} &
      &
      \multicolumn{1}{c|}{} &
      \multicolumn{1}{c|}{DaGAN ~\cite{hong2022depth}} &
      1800 (V) \\ \cline{9-10} 
      &
      &
      \multicolumn{2}{c|}{} &
      &
      \multicolumn{1}{c|}{} &
      &
      \multicolumn{1}{c|}{} &
      \multicolumn{1}{c|}{OneShotTH ~\cite{wang2021one}} &
      1800 (V) \\ \cline{8-10} 
      &
      &
      \multicolumn{2}{c|}{} &
      &
      \multicolumn{1}{c|}{} &
      &
      \multicolumn{1}{c|}{\multirow{2}{*}{summary}} &
      \multicolumn{1}{c|}{\multirow{2}{*}{\textbf{12}}} &
      \multirow{2}{*}{\textbf{21506 (V)}} \\
      &
      &
      \multicolumn{2}{c|}{} &
      &
      \multicolumn{1}{c|}{} &
      &
      \multicolumn{1}{c|}{} &
      \multicolumn{1}{c|}{} &
      \\ \hline
    \end{tabular}
  }
	\caption {Comparison of our multimodal facial attack datasets with Grandfake and JFSFDB. Our UniAttackData dataset covers all attack types, containing advanced forgery methods from 2020 to 2023, using the same ID as the existing dataset CASIA-SURF CeFA. Images in UniAttackData of the amount of $2,526,432$ are 3 times more than GrandFake. [Keys: I=Image, V=Video]}
	\label{tab:datasets}
\end{table*}

\subsection{Physical-Digital Attack Detection}
In a recently published paper, Yu et al ~\cite{yu2022benchmarking} established the first joint benchmark for face fraud and forgery detection and combined visual appearance and physiological rPPG signals to alleviate the generalization problem. Debayan Deb ~\cite{deb2023unified} et al. classified all 25 attack types documented in the literature and proposed a method to distinguish between real identities and various attacks using a multi-task learning framework and k-means enhancement techniques. However, none of these works have investigated unified attack detection based on ID consistency. 

\section{UniAttackData Dataset}
\subsection{Acquisition Detail} 
\begin{figure*}[ht]
	\centering
	\includegraphics[width=1\linewidth]{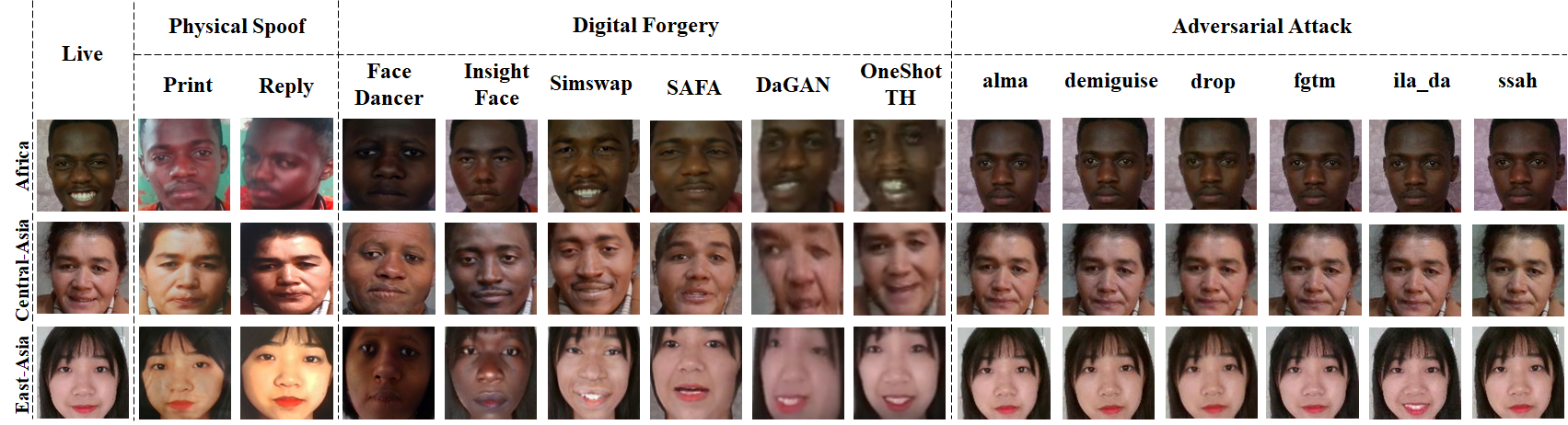}
	\caption{UniAttackData Dataset examples of all attack types corresponding to the same face ID. From top to bottom, they are Africans, Central Asians, and East Asians, respectively. The attack type of each sample is marked at the top.}
	\label{fig:enter-label}
\end{figure*}

As depicted in Tab.~\ref{tab:datasets}, our UniAttackData is an extension of CASIA-SURF CeFA~\cite{liu2021casia} through digital forgery, which includes $1,800$ subjects from $3$ ethnicities (e.g., African, East Asian and Central Asian), and $2$ types of physical attacks (e.g., Print and Replay). For each subject, we forge $12$ types of digital attacks as follows: (1) Pairing each subject video with a video of others as the reference video. (2) For digital forging, we employ the latest $6$ digital editing algorithms and $6$ adversarial algorithms on each live video, as detailed in Tab.~\ref{tab:datasets}.
(3) To selectively forge only the facial region while preserving the background, our process initiates face detection on every frame of the video. Subsequently, we digitally manipulate solely the facial area and seamlessly integrate the background onto the altered face. In overview, as depicted in Fig.~\ref{fig:enter-label}, our UniAttackData contains live faces from three ethnicities, two types of print attacks in distinct environments, one playback attack, and $6$ digital editing attacks and $6$ adversarial attacks.

\subsection{Advantages of UniAttackData.} In comparison to previously datasets, such as GrandFake~\cite{deb2023unified} and JFSFDB~\cite{yu2022benchmarking}, the UniAttackData has the following advantages: \textbf{Advantage 1.} Each ID contains a complete set of attack types. Because we conduct comprehensive physical and digital attacks on each live face, each ID contains a complete set of attack types. On the other hand, GrandFake and JFSFDB merge existing physical and digital attack datasets through integration, resulting in incomplete attack types for each ID, which easily leads to overfitting of the model to identity information. \textbf{Advantage 2.} Incorporation of the most advanced and comprehensive attack methods. Our UniAttackData contains $6$ editing attacks and $6$ adversarial attacks, utilizing algorithms developed from 2020 onwards for digital attack sample production. In contrast, JFSFDB only contains four types of digital editing attacks, and GrandFake uses algorithms predating 2020 for digital forgery. \textbf{Advantage 3.} The largest joint physical and digital attack dataset. As illustrated in Tab.~\ref{tab:datasets}, UniAttackData stands out as the dataset with the highest video count. Statistically, it comprises a total of $29,706$ videos, encompassing $1,800$ videos featuring live faces, $6,400$ videos showcasing physical attacks, and $21,506$ videos with digital attacks.

\subsection{Protocols and Statistics}
We define two protocols for UniAttackData to fully evaluate the performance of unified attack detection. (1) Protocol 1 aims to evaluate under the unified attack detection task. Unlike the classical single-class attack detection, the unified attack data protocol contains both physical and digital attacks. The huge intra-class distance and diverse attacks bring more challenges to the algorithm design. As shown in Tab.~\ref{tab:protocol}, the training, validation, and test sets contain live faces and all attacks. (2) Protocol 2 evaluates the generalization to ``unseen'' attack types. The large differences and unpredictability between physical-digital attacks pose a challenge to the portability of the algorithms. In this paper, we use the ``leave-one-type-out testing'' approach to divide Protocol 2 into two sub-protocols, where the test set for each self-sub-protocol is an unseen attack type. 
As shown in Tab~\ref{tab:protocol}, the test set of protocol 2.1 contains only physical attacks that have not been seen in the training and development sets, and the test set of protocol 2.2 contains only digital attacks that have not been seen in the training and development set.
\begin{table}[]
	\centering
	\scalebox{0.8}{
		\begin{tabular}{|c|c|cccc|c|}
			\hline
			\multirow{2}{*}{Protocol} & \multirow{2}{*}{Class} & \multicolumn{4}{c|}{Types}                                                                   & \multirow{2}{*}{\# Total} \\ \cline{3-6}
			&                        & \multicolumn{1}{c|}{\# Live} & \multicolumn{1}{c|}{\# Phys} & \multicolumn{1}{c|}{\# Adv}   & \#Digital &                        \\ \hline
			\multirow{3}{*}{P1}       & train                  & \multicolumn{1}{c|}{3000} & \multicolumn{1}{c|}{1800} & \multicolumn{1}{c|}{1800}  & 1800    & 8400                   \\ \cline{2-7} 
			& eval                   & \multicolumn{1}{c|}{1500} & \multicolumn{1}{c|}{900}  & \multicolumn{1}{c|}{1800}  & 1800    & 6000                   \\ \cline{2-7} 
			& test                   & \multicolumn{1}{c|}{4500} & \multicolumn{1}{c|}{2700} & \multicolumn{1}{c|}{7106}  & 7200    & 21506                  \\ \hline
			\multirow{3}{*}{P2.1}     & train                  & \multicolumn{1}{c|}{3000} & \multicolumn{1}{c|}{0}    & \multicolumn{1}{c|}{9000}  & 9000    & 21000                  \\ \cline{2-7} 
			& eval                   & \multicolumn{1}{c|}{1500} & \multicolumn{1}{c|}{0}    & \multicolumn{1}{c|}{1706}  & 1800    & 5006                   \\ \cline{2-7} 
			& test                   & \multicolumn{1}{c|}{4500} & \multicolumn{1}{c|}{5400} & \multicolumn{1}{c|}{0}     & 0       & 9900                   \\ \hline
			\multirow{3}{*}{P2.2}     & train                  & \multicolumn{1}{c|}{3000} & \multicolumn{1}{c|}{2700} & \multicolumn{1}{c|}{0}     & 0       & 5700                   \\ \cline{2-7} 
			& eval                   & \multicolumn{1}{c|}{1500} & \multicolumn{1}{c|}{2700} & \multicolumn{1}{c|}{0}     & 0       & 4200                   \\ \cline{2-7} 
			& test                   & \multicolumn{1}{c|}{4500} & \multicolumn{1}{c|}{0}    & \multicolumn{1}{c|}{10706} & 10800   & 26006                  \\ \hline
		\end{tabular}
	}
	\caption{Amount of train/eval/test images of different types under three different protocols: P1, P2.1, and P2.2.}
	\label{tab:protocol}
\end{table}
\section{Proposed Method}
\begin{figure*}[ht]
	\centering
	\includegraphics[width=1\linewidth,height=0.5\textwidth]{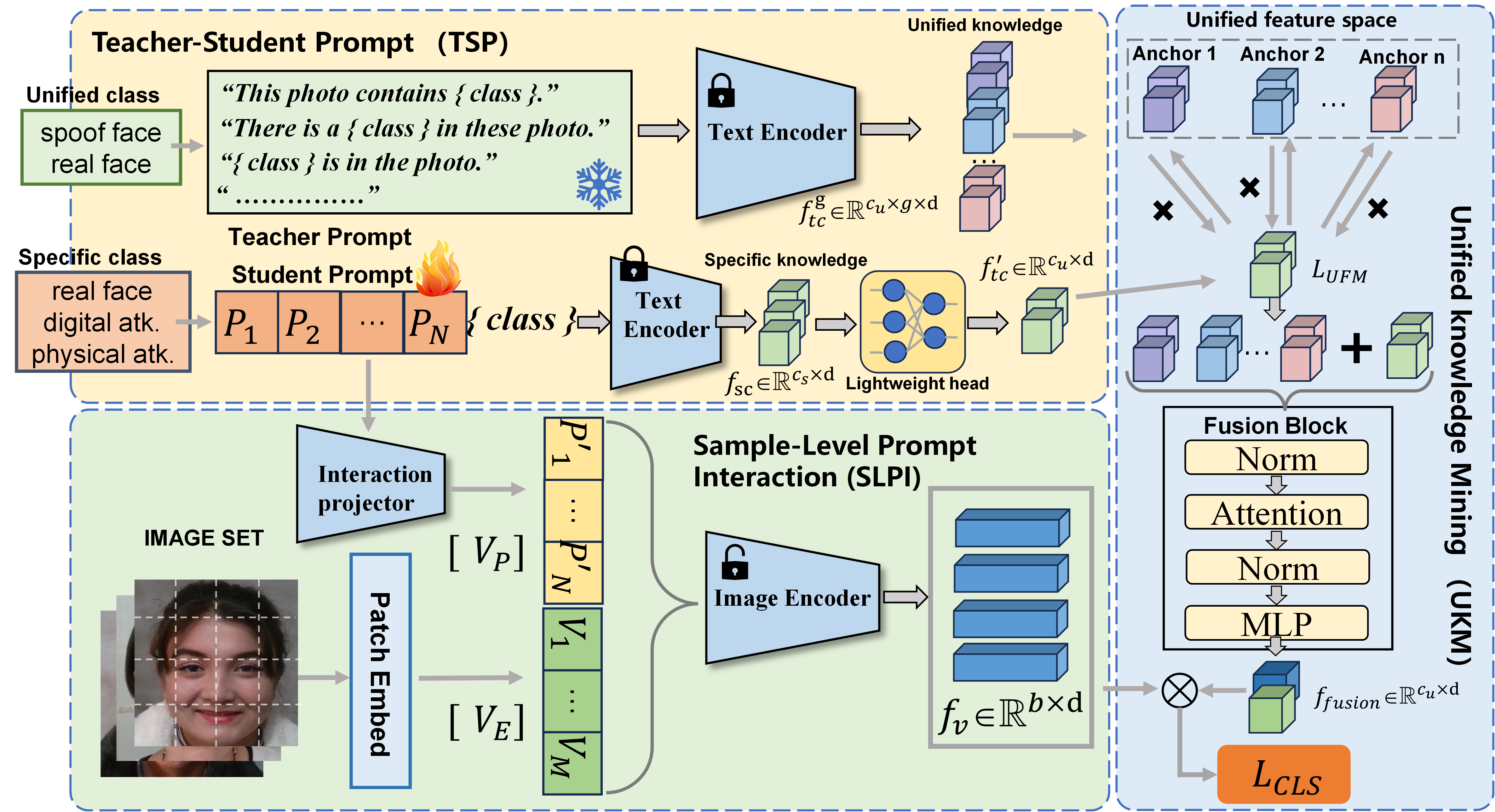}
	\vspace{-0.1cm}
	\caption{Our proposed UniAttackDetection architecture. The TSP module extracts unified and specific knowledge by constructing multiple groups of teacher prompts and learnable student prompts. The UKM module oversees the learning process by employing the unified knowledge mining loss, thereby enabling the model to acquire comprehensive insights across the entire feature space. The SLPI module maps the student prompts to the visual embedding space, allowing multi-modal prompt learning by making the student prompt learn sample-level semantics while allowing visual feature extraction to be guided by text.}
	\label{fig:network}
\end{figure*}


\subsection{Overview}
Our UniAttackDetection framework is based on CLIP~\cite{radford2021learning}. As illustrated in Fig.~\ref{fig:network}, the backbone network incorporates three key components: (1) The Teacher-Student Prompt (TSP) module, designed to extract unified and specific knowledge features by encoding both teacher prompts and student prompts, respectively; (2) The Unified Knowledge Mining (UKM) module, leveraging the Unified Feature Mining (UFM) loss to steer the student features, and facilitating the exploration of a tightly complete feature space. (3) The Sample-Level Prompt Interaction (SLPI) module, which maps learnable student prompts to visual modality prompts through the interaction projector. Subsequently, the language modality prompts obtained are linked to the original visual modality embeddings, and channel interactions take place during multi-modal prompt learning to extract sample-level semantics.

\subsection{Teacher-Student Prompt}
{\textbf{Teacher Prompt.}} Traditional FAS methods~\cite{wang2022domain} focus on extracting features commonly present in each category of images, such as facial contours and shapes. These features are then utilized to learn generic classification knowledge.
We introduce this concept into the natural language field by utilizing template prompts to guide the extraction of unified knowledge in physical-digital attacks, which is based on the following fact: The textual descriptions for each category can be effectively represented using a consistent hard template, implying a shared semantic feature space among them. Specifically, given a set of categories consisting of unified class names $unified_c$, $c$ ${\in}$ $\{ live face, spoof face \}$, the teacher prompt group is then obtained by interconnecting the class name and the manually designed template, i.e., $t_c$ = \textit{a photo of a} $\{unified_c\}$. This group of prompts is passed through CLIP's text encoder $G_T(\cdot)$ to compute the unified knowledge features of the text $f_{tc} = G_T(t_c)$, $f_{tc}$ ${\in}$ ${\mathbb{R}}^{c_u \times d}$, where $d$ is the dimension of CLIP. In this paper, we design multi-group artificial templates to fix discrete teacher anchors in the complete feature space. The multi-group teacher prompts are accumulated over the group dimension $g$ and finally encoded as $f_{tc}^g$ ${\in}$ ${\mathbb{R}}^{c_u \times g \times d}$.

\noindent{\textbf{Student Prompt.}} In contrast to conventional single-attack detection tasks, the Unified Attack Detection (UAD) task presents substantial gaps within the same classes. 
To mitigate this problem, we construct a group of student prompts. Student prompts are learned by minimizing the classification error on a training set consisting of specific classes, and thus student prompts possess strong learning abilities in specific classes. Specifically, we prefix each specific class $specific_c$, $c$ ${\in}$ $\{ real face, digital attack, physical attack \}$ with a series of learnable vectors $p_n \in {\mathbb{R}}^{d}, n \in {1,2,3,.... ,N}$ to get the student prompt. Then the student feature $f_{sc} = G_T(s_c)$ for learning specific knowledge is obtained, where $f_{sc}$ ${\in}$ ${\mathbb{R}}^{c_s \times d}$. To use the learned continuous tight features for binary classification, a lightweight head $ \mathcal{H}: {\mathbb{R}}^{c_s \times d} \rightarrow {\mathbb{R}}^{c_u \times d} $ is adopted to map specific knowledge to a generic classification space. 

\subsection{Unified Knowledge Mining}
To enhance the model's exploration of the complete feature space, we design the Unified Feature Mining (UFM) loss as follows:
\begin{equation}
	\mathcal{L}_{UFM}=\frac{1}{C} \sum_{g=0}^{G-1}\sum_{i=0}^{C-1}\left(1-\frac{f_{fusion} {f_{tc}^g}}{\left\|f_{fusion} ||\right\| {f_{tc}^g} \|}\right)
	\label{eq:HSR}
\end{equation}
This loss ensures the student features are sufficiently close to each teacher anchor in the unified knowledge feature space, yielding the following advantages: (1) The student prompt is no longer over-fitted with a specific knowledge distribution through the guidance of teacher anchors, and thus gains strong generalization capabilities. (2) Taking the learnable prompt as a carrier, the unlearnable unified feature space is fully mined to learn the tight and complete feature space.

Not only that, VLMs pre-trained on large-scale image-text pairs encapsulate abundant statistical common knowledge, which leads to the inclusion of potential semantic associations among text features extracted based on these models. Therefore, to capture the correlation between unified and specific knowledge features, we employ a fusion block to extract fusion features for multiple groups of teacher prompts and student prompts. 
Specifically, we concatenate multiple groups of teacher features $f_{tc}^g$ and student features into complete features $f_{com} \in {\mathbb{R}}^{c_u \times (g+1)  \times d }$ in the dimension of $g$. The fusion block consists of a self-attentive encoder $Att(\cdot)$ and a Multi-Layer Perceptron (MLP) $MLP_{fusion}(\cdot)$, where the self-attention encoder is given by the following equation:
\begin{equation}
	Att(X) = \frac{softmax(W_QX)(W_KX)^T}{\sqrt{d_k}}(W_VX)
	\label{eq:ATT}
\end{equation}
Where $W_Q$, $W_K$, and $W_V$ are the weight matrix of the query, key, and value, respectively, and $d_k$ is the dimension of the model. Finally, the modulated fusion knowledge feature ${f_{fusion}}$ is obtained by:
\begin{equation}
	{f_{fusion}} =MLP_{fusion}(Att(f_{com}))
	\label{eq:UKB}
\end{equation}

\subsection{Sample Level Prompt Interaction}
In recent FAS work~\cite{zhou2023instance}, learned sample-level semantics of visual pictures contribute to enhancing model robustness. Therefore, in VLMs, we deem it essential to establish a multi-modal interaction between picture instance perception features and student prompts. Student prompts can dynamically understand the sample-level semantics in visual pictures and thus learn comprehensive adaptive features. In turn, the visual modality will be guided by the shared prompts to deeply mine the category-related picture features. Inspired by MaPLe~\cite{khattak2023maple}, we design an interaction projector $p = Linear(\cdot)$. This interaction projector is implemented as a linear layer that maps the dimensions $d_p$ of student prompt into the  $d_v$ of visual embeddings. 
CLIP's text encoder takes the learnable token $[p_1][p_2][p_3] \dots [p_n]$ projected into word embeddings $[P]$ ${\in}$ ${\mathbb{R}}^{n \times d_p}$. The interaction projector then acts as a bridge to map the student prompt dimension word embeddings $[P]$ into visual dimension embeddings $[V_P]$, where the projector is expressed by a linear layer. The overall formulation is as follows:
\begin{equation}
	[V_P] = Linear([P]), [V] {\in} {\mathbb{R}}^{n \times d_v}
	\label{eq:stc}
\end{equation}
At the same time, the input image is encoded by the CLIP's patch embed layer as patch embeddings $[V_E]$ ${\in}$ ${\in}$ ${\mathbb{R}}^{m \times d_v}$, where $m$ is the patch dimension of the CLIP. Then, $[V_E]$ and $[V_P]$ are concatenated as the final visual embedding $V = [V_E, V_P] $, where $[\cdot, \cdot]$ refers to the concatenation operation. Finally, the visual embedding is passed through the CLIP image encoder $G_V$ to get the visual features $f_{v} = G_v(V)$. During the training process, we compute the cosine similarity between the image features and the labeled features to get the prediction probability of the image for each category and finally use the cross-entropy loss $L_{CLS}$ for binary class supervision. Together with the Unified Feature Mining (UFM) loss, the final objective is defined as:
\begin{equation}
	L_{Total} = L_{CLS} + \lambda \cdot L_{UFM}
	\label{eq:lossall}
\end{equation}
where $\lambda$ is a hyper-parameter to trade-off between two losses.

\section{Experiments}
\subsection{Experimental Setting}
{\textbf{Datasets.}} To evaluate the performance of the proposed method and existing approaches, we employ four datasets for face forgery detection, i.e., Our proposed UniAttackData, FaceForensics++ (FF++)~\cite{rossler2019faceforensics++}, OULU-NPU~\cite{boulkenafet2017oulu} and JFSFDB~\cite{yu2022benchmarking}. 
Our approach first demonstrates the superiority of our method on UniAttackData. Subsequently, we conducted sufficient experiments on other UAD datasets, such as the OULU-FF data protocol composed of OULU-NPU and FF++, and the JFSFDB dataset, to further demonstrate the generalization capability of our approach. Finally, our ablation study demonstrates the effectiveness of each component of our approach.

\noindent{\textbf{Evaluation metrics.}} For a comprehensive measure of the algorithm's UAD performance, we adopt the common metrics used in both physical forgery detection and digital forgery detection work. All experiments were conducted using average classification error rate ACER, overall detection accuracy ACC, the area under the curve (AUC), and equivalent error rate (EER) for performance evaluation. ACER and ACC on each test set are determined by the performance thresholds on the development set.

\noindent{\textbf{Implementation Details.}}
To better demonstrate the effectiveness of our method, we consider a series of existing competitors in the field of face anti-spoofing and base network backbone. The considered comparison methods are the ResNet50~\cite{he2016deep}, ViT-B/16~\cite{dosovitskiy2020image}, FFD~\cite{dang2020detection}, CDCN~\cite{yu2020searching} and the Auxiliary(Depth)~\cite{Liu2018Learning}. 
In addition, we incorporated our baseline approach into the comparative analysis. Notably, the student prompts for the baseline method were constructed through the unified class. In this way, we validate the superiority of our overall framework and the effectiveness of learning a continuous and compact specific class space.

\subsection{Experiments on Proposed UniAttackData}
\setlength{\tabcolsep}{10mm}{
\begin{table}[ht]
\renewcommand\arraystretch{1} 
\centering
\setlength{\tabcolsep}{1mm}
\scalebox{0.63}{ 
\begin{tabular}{c|c|cccc}

\toprule[1pt] 
\textbf{Prot.}     & \textbf{Method}         & \textbf{ACER(\%)$\downarrow$}    & \textbf{ACC(\%)$\uparrow$}     & \textbf{AUC(\%)$\uparrow$}    & \textbf{EER(\%)$\downarrow$}    \\ \midrule[1pt]
\multirow{5}{*}{\textbf{1}} & ResNet50                & 1.35                 & 98.83                & 99.79               & 1.18                \\
                   & VIT-B/16                & 5.92                 & 92.29                & 97.00               & 9.14                \\
                   & Auxiliary               & 1.13                 & 98.68                & 99.82               & 1.23              \\
                   & CDCN                    & 1.40                 & 98.57                & 99.52               & 1.42                \\
                   & FFD                     & 2.01                 & 97.97                & 99.57                & 2.01                \\
                   & Baseline(Our)                     & 0.79                 & 99.44                & 99.96                & 0.53                \\
                    \rowcolor{mygray}
                   & \textbf{UniAttackDetection(Our)} & \textbf{0.52}        & \textbf{99.45}       & \textbf{99.96}      & \textbf{0.53}       \\ \midrule[1pt]
\multirow{5}{*}{\textbf{2}} & ResNet50                & 34.60$\pm$5.31           & 53.69$\pm$6.39           & 87.89$\pm$6.11          & 19.48$\pm$9.10         \\
                   & VIT-B/16                & 33.69$\pm$9.33           & 52.43$\pm$25.88          & 83.77$\pm$2.35          & 25.94$\pm$ 0.88          \\
                   & Auxiliary               & 42.98$\pm$6.77           & 37.71$\pm$26.45          & 76.27$\pm$12.06         & 32.66$\pm$7.91          \\
                   & CDCN                    & 34.33$\pm$0.66           & 53.10$\pm$12.70          & 77.46$\pm$17.56         & 29.17$\pm$14.47         \\
                   & FFD                     & 44.20$\pm$1.32            & 40.43$\pm$14.88         & 80.97$\pm$2.86          & 26.18$\pm$2.77               \\
                   & Baseline(Our)                     &28.90$\pm$ 10.85            & 58.28$\pm$ 26.37         & 89.48$\pm$ 4.94          & 19.04$\pm$ 5.81               \\
                   \rowcolor{mygray}
                   & \textbf{UniAttackDetection(Our)} & \textbf{22.42$\pm$ 10.57} & \textbf{67.35$\pm$ 23.22} & \textbf{91.97$\pm$ 4.55} & \textbf{15.72$\pm$ 3.08} \\ \bottomrule[1pt]
\end{tabular}
}
\caption{The results of intra-testing on two protocols of UniAttackData, where the performance of Protocol 2 quantified as the mean$\pm$std measure derived from Protocol 2.1 and Protocol 2.2.}
\label{Table:p1}
\end{table}
}

\noindent{\textbf{Experiments on Protocol 1.}} In protocol $1$ of UniAttackData, the data distribution is relatively similar across sets. The training, development, and test sets all contain both physical and digital attacks. This protocol is suitable for evaluating the performance of the algorithms in UAD tasks. We present the performance results for commonly used backbone networks, networks for physical attack detection, and networks for digital attack detection. As shown in Tab~\ref{Table:p1}, our algorithm outperforms others in all four metrics—ACER, ACC, AUC, and EER. This proves the superiority of our method for unified attack detection.

\noindent{\textbf{Experiments on Protocol 2.}}  
In protocol $2$ of UniAttackData, the test set comprises attack forms that are ``unseen'' in the training or validation sets, aiming to assess the algorithm's generalizability. As shown in Tab~\ref{Table:p1}, an interesting observation emerges: the performance of single-attack detection networks FFD, CDCN, and Auxiliary largely degrades on Protocol 2, and almost all of their four metrics are lower than classic backbone ViT and ResNet, which exposes the flaw that the single-attack detector is over-fitting specific attacks. On the other hand, our method ranks first in performance on all metrics, which proves that our method not only learns unified knowledge of the UAD task but also captures a compact and continuous feature space for attack categories.

\subsection{Experiments on Other UAD Datasets}

\setlength{\tabcolsep}{10mm}{
\begin{table}[ht]
\renewcommand\arraystretch{1} 
\centering
\setlength{\tabcolsep}{1mm}
\scalebox{0.70}{ 
\begin{tabular}{c|c|cccc}
      \toprule[1pt] 
      \textbf{Data.} & \textbf{Method} & \textbf{ACER(\%)$\downarrow$} & \textbf{ACC(\%)$\uparrow$} & \textbf{AUC(\%)$\uparrow$} & \textbf{EER(\%)$\downarrow$} \\
      \midrule[1pt]
      & ResNet50 & 7.70 & 90.43 & 98.04 & 6.71 \\
      & VIT-B/16 & 8.75 & 90.11 & 98.16 & 7.54 \\
      & Auxiliary & 11.16 & 87.40 & 97.39 & 9.16 \\
      & CDCN & 12.31 & 86.18 & 95.93 & 10.29 \\
      & FFD & 9.86 & 89.41 & 95.48 & 9.98 \\
      & Baseline(Our) & 2.84 & 96.93 & 99.54 & 2.90 \\
      \multirow{-6}{*}{JFS-FDB} & \cellcolor{mygray}\textbf{UniAttackDetection(Our)} & \cellcolor{mygray}\textbf{1.66} & \cellcolor{mygray}\textbf{98.23} & \cellcolor{mygray}\textbf{99.74} & \cellcolor{mygray}\textbf{1.78} \\
      \midrule[1pt]
      & ResNet50 & 7.45 & 92.64 & 96.92 & 7.60 \\
      & VIT-B/16 & 9.95 & 90.44 & 97.30 & 9.83 \\
      & Auxiliary & 16.43 & 83.68 & 92.98 & 16.80 \\
      & CDCN & 17.42 & 82.56 & 92.37 & 17.35 \\
      & FFD & 19.13 & 80.32 & 88.90 & 19.00 \\
      & Baseline(Our) & 2.22 & 97.52 & 99.60 & 2.22 \\
      \multirow{-6}{*}{OULU-FF} & \cellcolor{mygray}\textbf{UniAttackDetection(Our)} & \cellcolor{mygray}\textbf{1.63} & \cellcolor{mygray}\textbf{98.00} & \cellcolor{mygray}\textbf{99.81} & \cellcolor{mygray}\textbf{1.80} \\
      \bottomrule[1pt] 
    \end{tabular}
}
\caption{The results of Unified attack data protocol JFSFDB and OULU-FF.}
\label{Table:ffoulu}
\end{table}
}

To verify the superiority of our proposed dataset and algorithm in more detail. We have done extensive experiments on other UAD datasets. As shown in Tab~\ref{tab:protocol}, we obtained the best performance for all metrics on the previously proposed JFSFDB dataset. Furthermore, we combine the training, validation, and testing sets provided by FF++~\cite{rossler2019faceforensics++} and OULU-NPU~\cite{boulkenafet2017oulu} to form a unified attack data protocol named OULU-FF. Our algorithm also outperformed competing methods on the OULU-FF protocol. This validates the ability of our algorithm to detect unified attacks. Notably, we observed a significant decline in the performance of all evaluated methods on both joint attack datasets in comparison to our proposed dataset. This finding underscores the challenge algorithms face in discerning spoofing traces without ID consistency.

\subsection{Ablation Experiment}
\setlength{\tabcolsep}{10mm}{
\begin{table}[ht]
\renewcommand\arraystretch{1} 
\centering
\setlength{\tabcolsep}{1mm}
\scalebox{0.8}{ 
\begin{tabular}{c|cccc}
\toprule[1pt] 
\textbf{Method}           & \textbf{ACER(\%)}$\downarrow$ & \textbf{ACC(\%)}$\uparrow$ & \textbf{AUC(\%)$\uparrow$} & \textbf{EER(\%)}$\downarrow$ \\ \midrule[1pt]
UniAttackDetection w/o $s_c$        & 1.76              & 96.95            & 99.90            & 2.03             \\
UniAttackDetection w/o $t_c$        & 0.83              & 99.40            & 99.95            & 0.60             \\
UniAttackDetection w/o $L_{UKM}$      &0.71              & 99.25            & 99.95            & 0.71             \\
UniAttackDetection w/o $V_P$        & 2.56              & 98.70            & 99.90            & 0.93             \\
\rowcolor{mygray}
\textbf{UniAttackDetection(Our)}   &\textbf{0.52}      & \textbf{99.45}   & \textbf{99.95}   & \textbf{0.53}    \\ \bottomrule[1pt]
\end{tabular}
}
\caption{The ablation study of different Components. The evaluation protocol is P1.}
\label{Table:ablation-1}
\end{table}
}

\noindent{\textbf{Effectiveness of Different Components.}} To verify the superiority of our UniAttackDetection as well as the contributions of each component, multiple incomplete models are built up by controlling different variables. All results are measured in the same manner, as shown in Tab~\ref{Table:ablation-1}. First, to verify the validity of student prompts and teacher prompts, we conducted experiments without student prompts or teacher prompts, respectively. The experiments showed that both unified and specific knowledge play a positive role in the UAD task. In addition, to validate the effectiveness of the Unified Knowledge Mining (UKM) module, we conducted experiments with UniAttackDetection $w/o$  $L_{HSR}$. Specifically, we directly let the unified knowledge features and the specific knowledge features be connected as the final text features. The experiments demonstrate that the UKM module helps the model learn unified knowledge and enhances the generalization to different attacks. Then, to demonstrate the importance of the Sample-Level Prompt Interaction (SPLI) module, we conducted experiments without adding visual prompts. The quantitative results show that multi-modal prompt learning facilitates improved learning of instance-level features.

\begin{figure}[t]
\centering
\includegraphics[width=1.0\linewidth]{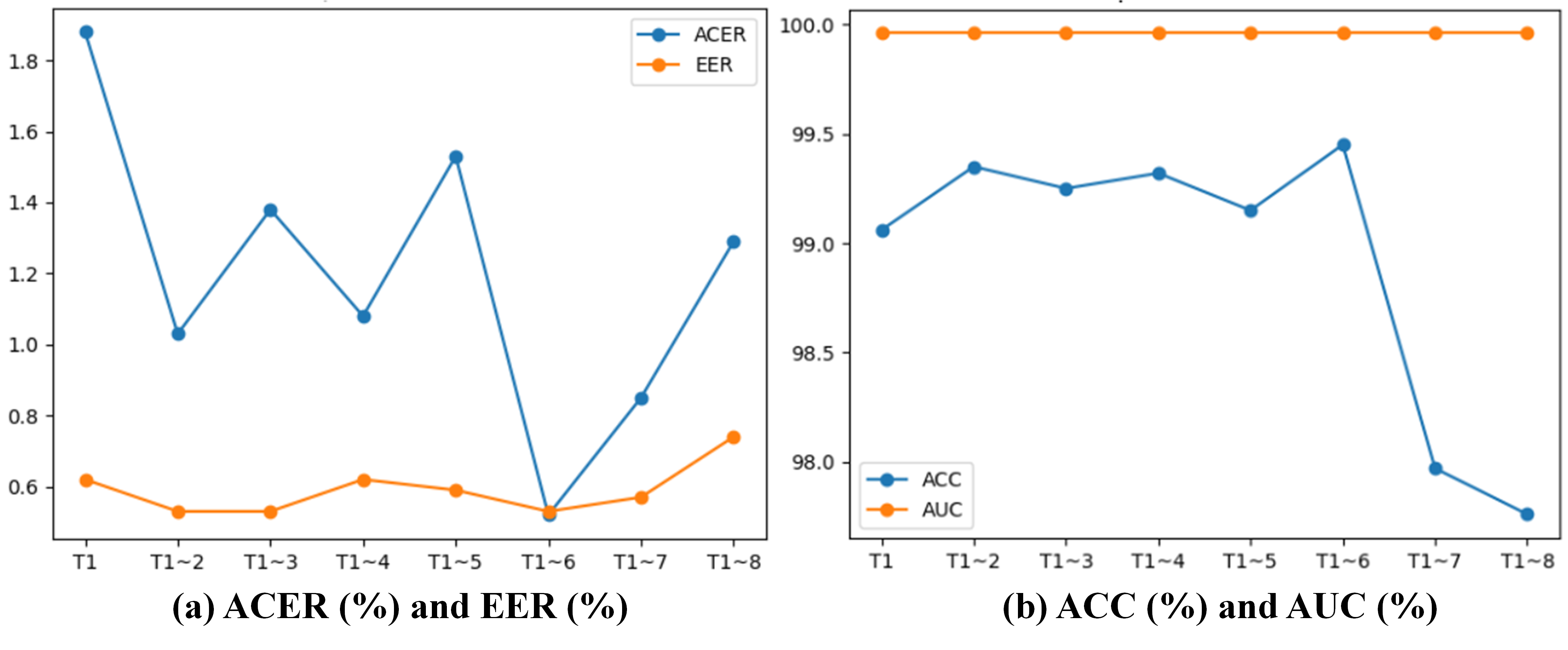}
\caption{Ablation experiments of the selection of teacher prompts. The horizontal coordinates indicate which teacher prompts from Tab~\ref{Table:ablation-2} were selected. For example, T1$\sim$6 indicates the selection of the first six prompts from Tab~\ref{Table:ablation-2}. }
\label{fig:ablation-3}
\end{figure}

\noindent{\textbf{Effectiveness of Teacher Prompts.}} 
To validate the effect of teacher prompts on experimental performance, we incrementally introduced new templates into the teacher prompts group. Tab~\ref{Table:ablation-2} shows the specific language descriptions covering the real and spoof categories. Fig~\ref{fig:ablation-3} depicts the variation in experimental performance as we systematically increased the number of teacher prompt groups from one to eight. Examination of the figure reveals that, with the exception of the AUC (Area Under the Curve) metric, there is a discernible trend where the performance across the other three metrics initially improves with the addition of more prompts, only to subsequently decline. All four experimental metrics obtain the best results when six templates are used as teacher prompts. This proves that learning the complete feature space through multiple sets of teacher anchors is beneficial for the UAD task, but too many teacher anchors may instead hinder the adaptive learning ability of student prompts.

\setlength{\tabcolsep}{10mm}{
\begin{table}[ht]
\renewcommand\arraystretch{1} 
\centering
\setlength{\tabcolsep}{1mm}
\scalebox{0.90}{ 
\begin{tabular}{c|c}
\toprule[1pt] 
\textbf{Prompt No.} & \textbf{Teacher Prompts (Templates)}                                         \\ \midrule[1pt]
T1          & This photo contains \{real face\}/ \{spoof face\}.                         \\ \hline
T2          & There is a \{real face\}/ \{spoof face\} in this photo.                    \\ \hline
T3          & \{real face\}/ \{spoof face\} is in this photo.                           \\ \hline
T4          & A photo of a \{real face\}/ \{spoof face\}.                                \\ \hline
T5          & \multicolumn{1}{l}{This is an example of a \{real face\}/ \{spoof face\}.} \\ \hline
T6          & This is how a \{real face\}/ \{spoof face\} looks like. \\ \hline
T7          & This is an image of \{real face\}/ \{spoof face\}. \\ \hline
T8          & The picture is a \{real face\}/ \{spoof face\}. \\ \bottomrule[1pt]
\end{tabular}
}
\caption{The multiple groups of manual templates used are fixed during training.}
\label{Table:ablation-2}
\end{table}
}

\subsection{Visualization Analysis}
\begin{figure}[t]
\centering
\includegraphics[width=1\linewidth]{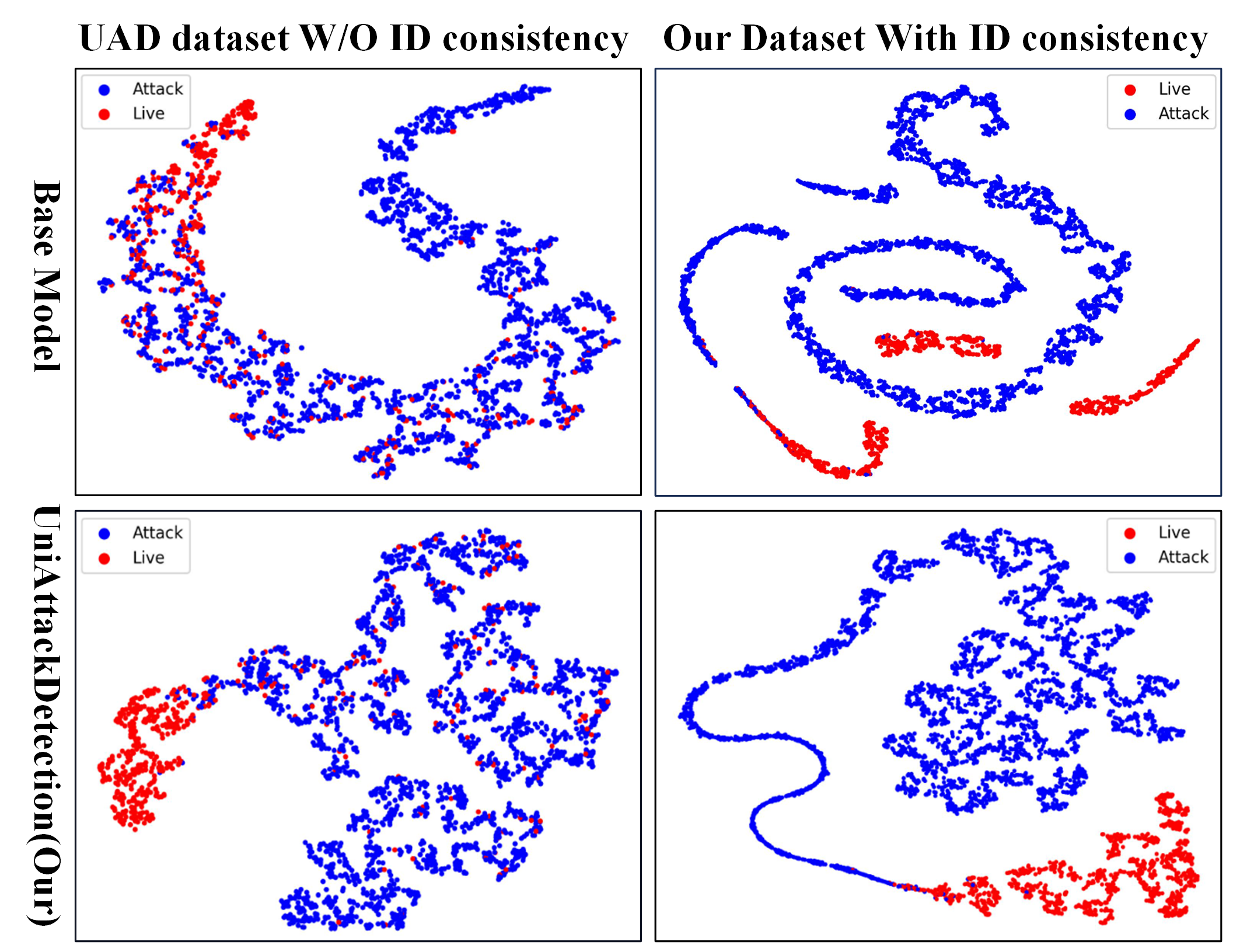}
\caption{Feature distribution comparison on UAD data protocol (OULU-FF) and UniAttackData using t-SNE. Different colors denote features from different classes.}
\label{fig:visiual}
\end{figure}
Here, we visualize the distribution of features learned by the baseline model ResNet50 and our method on UniAttackData's P1 and the previously mentioned UAD data protocol, respectively. As depicted in Fig~\ref{fig:visiual}, our study reveals two significant findings: (1) ID Consistency Impact. Both the base model ResNet50 and our proposed method exhibit limitations in classifying datasets lacking ID consistency assurance. The feature space displays a mixing of lives and attacks across various regions. In contrast, on the proposed dataset, both methods distinctly segregate lives from attacks. This underscores the critical role of ID consistency, enabling the model to concentrate on acquiring deception features rather than being influenced by ID-related noise features. (2) Enhanced Category Feature Space. In comparison to the baseline model, our approach demonstrates the capacity to learn a closely connected category feature space. This highlights our proficiency in acquiring category-specific knowledge. Moreover, our method successfully establishes a distinct category boundary on both datasets, affirming its effectiveness in extracting unified knowledge.
\section{Conclusion}
In this paper, we proposed a dataset that combines physical and digital attacks, called UniAttackData. This is the first unified attack dataset with guaranteed ID consistency. In addition, we proposed a unified detection framework based on CLIP, namely UniAttackDetection. The method introduces language information into the UAD task and substantially improves the performance of unified attack detection through multi-modal prompt learning. Finally, we conduct comprehensive experiments on UniAttackData and three other datasets to verify the importance of the datasets for the attack detection task and the effectiveness of the proposed method.

\bibliographystyle{named}
\bibliography{ijcai24}

\clearpage
\section*{Appendix}
\titleformat{\section}[display]{\bfseries\centering}{}{1em}{}

\section{CASIA-SURF CeFA}
\begin{figure*}[ht]
	\begin{center}
		\includegraphics[width=1.0\linewidth]{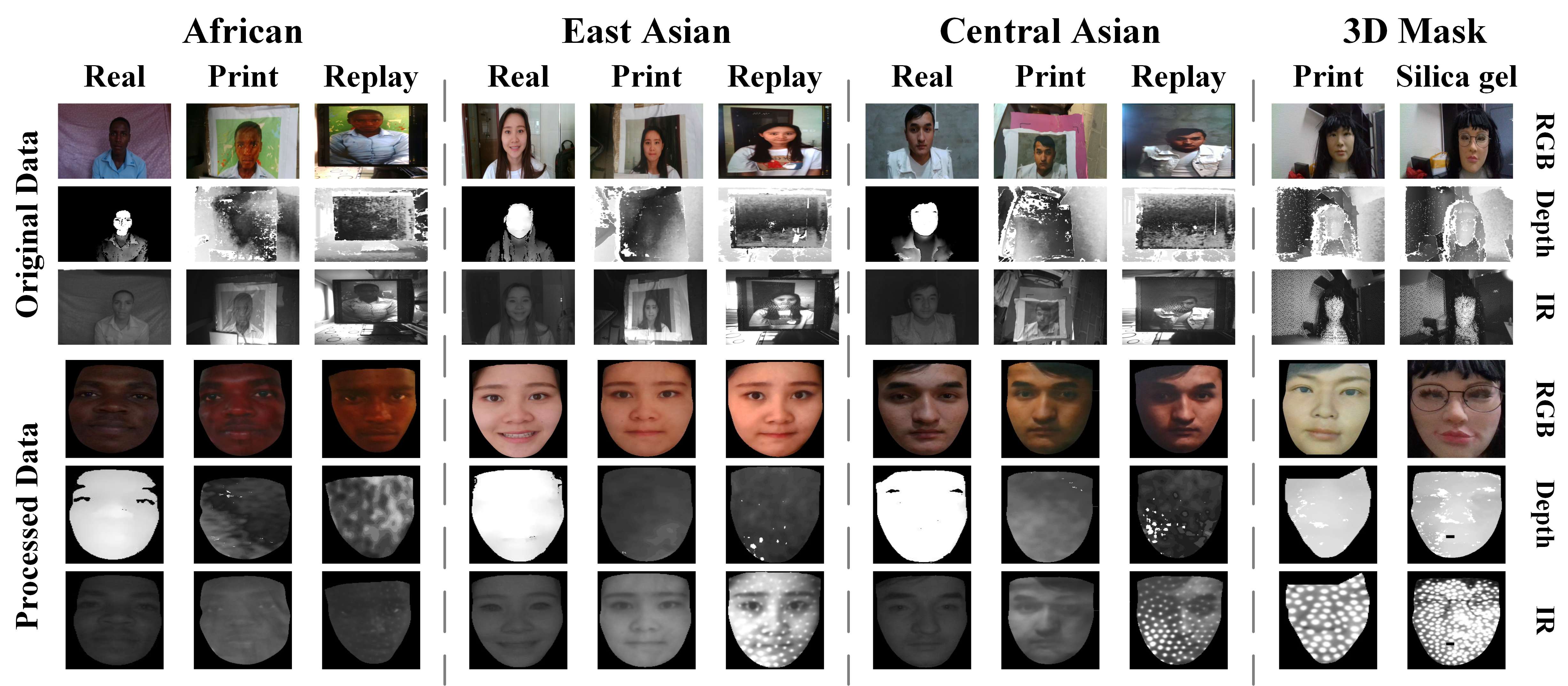}
	\end{center}
	\caption{Samples of the CASIA-SURF CeFA dataset. It contains $1,607$ subjects, $3$ different ethnicities (i.e., Africa, East Asia, and Central Asia), with $4$ attack types (i.e., print attack, replay attack, 3D print and silica gel attacks).}
	\label{samp_data}
\end{figure*}

To alleviate the ethnic bias and ensure that face PAD methods are in a safe reliable condition for users of different ethnicities, Liu~et.al~\cite{liu2021casia} introduced the largest up-to-date Cross-ethnicity Face Anti-spoofing (CeFA) dataset, covering $3$ ethnicities, $3$ modalities, $1,607$ subjects, and 2D plus 3D attack types. As shown in Fig.~\ref{samp_data}, CeFA consists of 2D and 3D attack subsets. The 2D attack subset, consists of print and video-replay attacks captured by subjects from three ethnicities (e.g., African, East Asian, and Central Asian). Each ethnicity has 500 subjects, and each subject has $1$ real sample, $2$ fake samples of print attack captured indoor and outdoor, and $1$ fake sample of video replay.  

In this work, we make the following changes based on CeFA to construct our UniAttackData dataset: 1) Remove depth and near-infrared modalities. Due to the current digital attack technology generally based on RGB samples for editing, our dataset only retains RGB modality. 2) Remove 3D attacks. Due to the fact that the 3D masks in CeFA were not cast according to real faces, unknown identities appeared. Therefore, we remove the 3D masks to maintain the integrity of each identity including the attacks. 3) Add 100 subjects for each ethnicity. To increase the diversity, we add 100 subjects to each ethnicity and collect live videos and physical attack videos using devices and environments similar to CeFA. Such as we use the Intel Realsense to capture the videos at 30f ps. The resolution is 1280 × 720 pixels for each frame in the video. Subjects are asked to move smoothly their head so as to have a maximum of around 300 deviations of head pose in relation to the frontal view. 4) Perform digital forgery based on each live video.

\section{Digital Attacks}
For digital attacks, we consider 6 types of editing attacks (ie. FaceDancer, InsightFace. SimSwap, SAFA, DaGAN, OneShotTH) and 6 types of adversarial attacks (ie. advdrop, alma, demiguise, fgtm, ila\_da, ssah), all proposed after 2020.

\subsection{Digital Forgery Methods}
\textbf{\textit{FaceDancer}}~\cite{rosberg2023facedancer} introduces a unique, single-stage approach for high-fidelity face swapping that effectively balances the need for preserving original facial attributes with the accurate transfer of new facial identities. Adaptive Feature Fusion Attention (AFFA) intelligently learns to combine attribute features with identity-conditioned features and Interpreted Feature Similarity Regularization (IFSR) preserves essential attributes while transferring the identity. The approach ensures high fidelity in identity transfer while maintaining the integrity of the original facial features.

\textbf{\textit{DaGAN}}~\cite{hong2022depth} is a Depth-aware Generative Adversarial Network designed for talking head video generation. Self-supervised Face Depth Learning Module automatically recovers dense 3D facial geometry from videos, Depth-guided Facial Keypoints Detection Module combines geometry from depth maps with appearance from images and Cross-Modal Attention Mechanism utilizes depth-aware attention to refine the motion field, focusing on fine-grained details of facial structure and movements. DaGAN shows significant improvements in generating realistic faces, especially under complex conditions like varied poses and expressions.

\textbf{\textit{OneShotTH}}~\cite{wang2021one} focuses on one-shot free-view neural talking-head synthesis. Specifically, Single Image Utilization efficiently synthesizes talking-head videos from just a single source image and a driving video and 3D Keypoint Representation incorporates a novel 3D keypoint representation that unsupervisedly separates identity-specific features from motion-related information. Also, the Bandwidth Efficiency method is designed to be bandwidth-efficient, making it highly suitable for real-time applications, Realistic Video Synthesis achieves high-quality synthesis of talking-head videos. OneShotTH represents a significant advancement in digital synthesis techniques, offering a practical and efficient solution for generating realistic talking-head videos in bandwidth-limited scenarios.

\textbf{\textit{SAFA}}~\cite{wang2021safa} (Structure Aware Face Animation) primarily focuses on understanding and integrating the scene structure of a face to enhance the animation quality. 1) Combining 3D Morphable Model (3DMM) with 2D Affine Motion Model, which allows for detailed modeling of the face's scene structure; 2) Contextual Attention Module for Image Inpainting effectively recovers the areas perceived as occluded in the source image; 3) Geometrically-Adaptive Denormalization Layer improves the quality of face appearance in the animations by incorporating 3D geometry information. SAFA's approach demonstrates excellence in generating realistic, pose-preserving, and identity-preserving face animations, particularly in scenarios involving large pose deviations and occlusions.

\textbf{\textit{SimSwap}}~\cite{chen2020simswap} is an innovative digital face-swapping framework focusing on high fidelity and generalization across arbitrary identities. The ID Injection Module (IIM) transfers the identity of a source face into the target face at the feature level, Weak Feature Matching Loss helps preserve the facial attributes of the target face by aligning the generated result with the target at a high semantic level. SimSwap excels in identity transfer while maintaining attributes of the target face, even in diverse and challenging conditions.

\textbf{\textit{InsightFace}}~\cite{2020Deep} project contains a Picsi.AI, which is an advanced tool for creating personalized portraits, utilizing the InsightFaceSwap Discord bot and Midjourney. It offers high realism in portrait creation and includes features like HiFidelity Mode for enhanced quality and Sharpness Options for added detail. The tool allows users to swap faces in images, apply effects like aging, and support GIF creation with improved quality and size limits. For detailed operations, users can utilize Discord Slash Commands for various functions like saving, setting, listing, and deleting identity names, as well as swapping faces in images.

\subsection{Adversarial Attacks}
\textbf{\textit{AdvDrop}}~\cite{duan2021advdrop} is a novel approach utilizing frequency domain differencing and adversarial attack techniques, aimed at creating adversarial examples by removing subtle details from images. The underlying principle involves transforming the original image from the spatial domain to the frequency domain, followed by quantization of the transformed image to reduce its frequency components. This quantization process effectively removes certain frequency components in the frequency domain to eliminate subtle details from the image. The method focuses on quantifying features of interest related to "image details" in the frequency domain, taking into account the natural insensitivity of the human eye to subtle image details. 

\textbf{\textit{ALMA}}~\cite{rony2021augmented} (Augmented Lagrangian Method for Adversarial Attacks) is a customized augmented Lagrangian method designed for generating adversarial samples. This algorithm achieves highly competitive performance by combining internal and external iterations of the Lagrangian with joint updates for perturbations and multipliers. The ALMA attack algorithm can be applied to generate adversarial samples for various distance metrics, including $\ell$1-norm, $\ell$2-norm, CIEDE2000, LPIPS, and SSIM. In the algorithm, the improvement of penalty parameters $\mu$ and $\rho$ is a crucial design choice. In the ALMA attack algorithm, the use of Exponential Moving Average (EMA) to smooth the value of $\mu$ is employed to reduce the optimization budget required and enhance the stability of the optimization process. 

\textbf{\textit{Demiguise}}~\cite{wang2021demiguise} Demiguise attack is an algorithm designed for adversarial attacks, primarily based on the optimization of perceptual similarity and the utilization of high-dimensional semantic information. The algorithm adjusts parameters such as the compression ratio of JPEG compression and the bit depth of binary filters to generate adversarial perturbations. However, even while maintaining reasonable image quality, the Demiguise-C\&W attack consistently achieves higher deception rates in most cases, attributed to its optimization leveraging perceptual similarity. In comparison to other simple metric methods that introduce significant noise at the edges of images, diminishing imperceptibility, Demiguise-C\&W explores rich semantic information in high-order structured representations through perceptual similarity, resulting in imperceptible perturbations. 

\textbf{\textit{SSAH}}~\cite{luo2022frequency} (Semantic Similarity Attack with Low-frequency Constraint) is an adversarial attack algorithm designed for deep learning models. The algorithm enhances attack transferability and model independence by exploiting semantic similarity in the feature space of attacked representations. In this method, the classifier is deceived by manipulating the feature representations, steering them away from benign samples and closer to those of adversarial and target (least similar) samples, without requiring specific knowledge of image categories. The algorithm assumes that higher-level representations imply differences and semantics in images, guiding perturbations towards semantic regions within the pixel space. SSAH also introduces a low-frequency constraint, enhancing the imperceptibility of the attack perturbation by restricting perturbations to low-frequency components.

\textbf{\textit{ILA-DA}}~\cite{yan2022ila} (Intermediate Level Attack-Data Augmentation) is an adversarial attack algorithm designed for deep learning models, aiming to enhance attack transferability by augmenting reference attacks and iteratively searching for data augmentations. The implementation of this algorithm consists of three main components. Firstly, automatic data augmentation enhances attack transferability by applying computationally inexpensive transformations to input images. Secondly, the backward adversarial update utilizes adversarial noise to create new augmented references, and attack interpolation forms new augmented references by interpolating images in the direction that leads to greater intermediate feature differences. Specifically, ILA-DA utilizes automatic and effective image transformations, efficient backward adversarial update techniques, and attack interpolation methods to create more transferable adversarial samples.

\textbf{\textit{FGTM}}~\cite{zou2022making} (Adam Iterative Fast Gradient Tanh Method) is an algorithm designed for generating adversarial samples. This method addresses limitations in the traditional Fast Gradient Sign Attack series concerning indistinguishability and transferability, aiming to create adversarial samples that are more challenging to identify and easier to transfer. AI-FGTM improves upon the limitations of the basic symbolic structure attack method by modifying the primary gradient processing steps. By substituting the tanh function for the symbolic function in gradient processing, adopting the Adam algorithm in place of momentum methods and gradient normalization, and applying Gaussian blurring with smaller filters, the design of AI-FGTM aims to preserve gradient information as much as possible while generating nearly imperceptible perturbations. Additionally, the method employs dynamic step sizes and smaller filters to further enhance the attack success rate. 

\section{Proposed Method}
\begin{figure}[t]
	\begin{center}
		\includegraphics[width=1.0\linewidth]{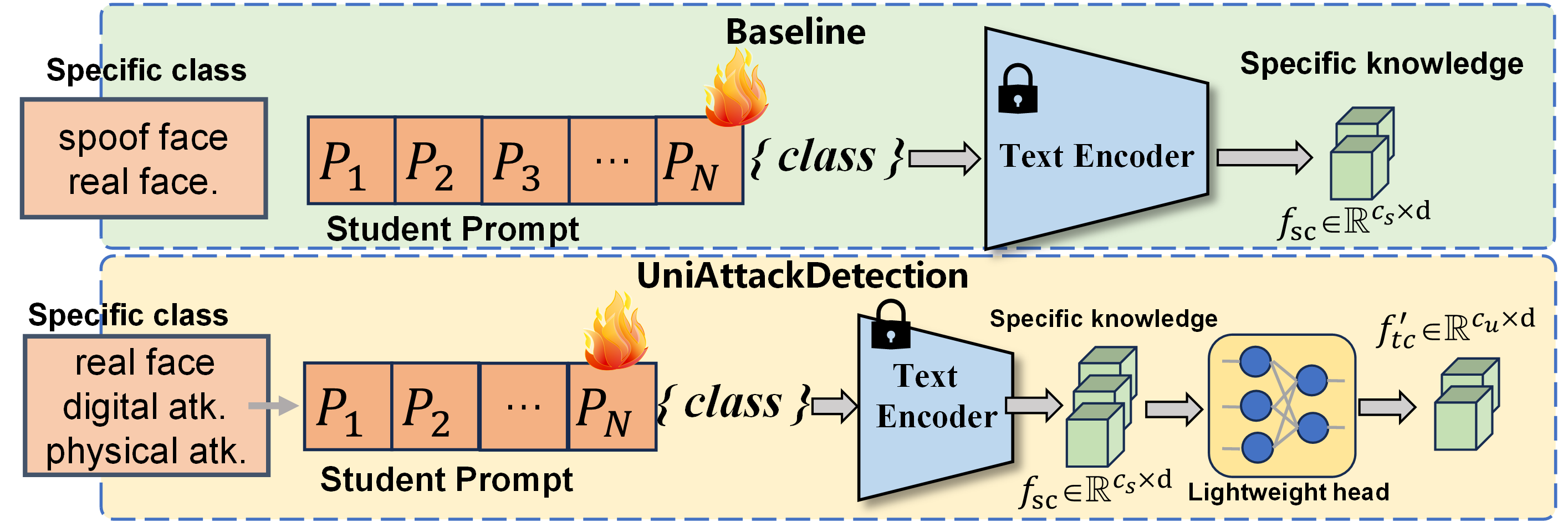}
	\end{center}
	\caption{Comparison of constructing student prompt module in the Baseline Framework and constructing student prompt module in UniAttackDetection.}
	\label{fig:appendix-main}
\end{figure}
\subsection{Architecture Comparison}
As shown in Fig~\ref{fig:appendix-main}, unlike the UniAttackDetection mentioned in the main article, we prefix each label $classname_c$, $c$ ${\in}$ $\{ live face, spoof face \}$ with a series of learnable vectors $p_n \in {\mathbb{R}}^{d}, n \in {1,2,3,.... ,N}$ to get the student prompt $s_c =  [p_1][p_2][p_3]... [p_n][classname_c]$. Finally, the student feature $f_{sc} = G_T(s_c)$ for learning category knowledge is obtained. Utilizing the powerful learning capabilities of student prompts, the baseline model allows for the learning of a broad category space. UniAttackDetection builds on this foundation by incorporating specific kinds of attacks into the category set $specific_c$,where $c$ ${\in}$ $\{ real \, face, digital \, attack, physical \, attack \}$. This construction of hints facilitates the model's learning of category-specific knowledge, leading to the learning of a tight and continuous category feature space. Additionally, a lightweight head was introduced, designed to map class-specific knowledge to a generic feature space for the eventual binary classification task.
\begin{figure}[h]
	\begin{center}
		\includegraphics[width=1.0\linewidth]{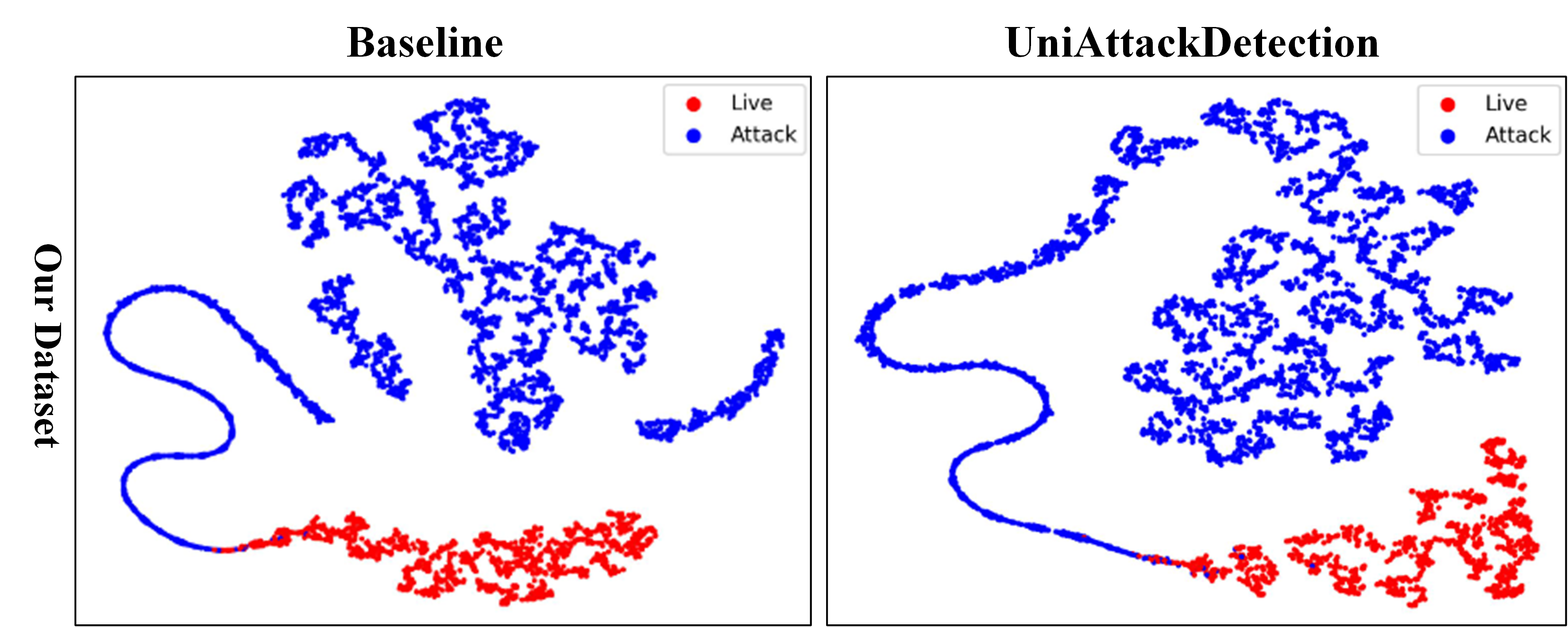}
	\end{center}
	\caption{Feature distribution comparison on protocol 1 of our dataset using t-SNE. Different colors denote features from different classes.}
	\label{fig:appendix-visual}
\end{figure}
\subsection{Visualisation Comparison}
Here, we visualize the baseline model mentioned above and the distribution of features learned by UniAttackDetection on protocol 1 of UniAttackData. As depicted in Fig.~\ref{fig:appendix-visual}, the baseline model exhibits commendable proficiency in discriminating between live and spoof, effectively establishing well-defined category boundaries. Nonetheless, the acquired feature space for attack categories appears to be loosely interconnected and discontinuous. In contrast, UniAttackDetection demonstrates robust learning of category spaces, consolidating all attack types cohesively without discernible gaps. This underscores the efficacy of acquiring category-specific knowledge and bolsters the model's performance in binary classification tasks through the implementation of lightweight heads.

\end{document}